# Automatic Cardiac Disease Assessment on cine-MRI via Time-Series Segmentation and Domain Specific Features

Fabian Isensee[1]⋆, Paul F. Jaeger[1]⋆, Peter M. Full[2,3], Ivo Wolf[3], Sandy Engelhardt[2,3], and Klaus H. Maier-Hein[1]

[1] Medical Image Computing, German Cancer Research Center (DKFZ), Heidelberg, Germany
[2] Division of Computer-assisted Medical Interventions, German Cancer Research Center (DKFZ), Heidelberg, Germany
[3] Department of Computer Science, Mannheim University of Applied Science, Mannheim, Germany

**Abstract.** Cardiac magnetic resonance imaging improves on diagnosis of cardiovascular diseases by providing images at high spatiotemporal resolution. Manual evaluation of these time-series, however, is expensive and prone to biased and non-reproducible outcomes. In this paper, we present a method that addresses named limitations by integrating segmentation *and* disease classification into a fully automatic processing pipeline. We use an ensemble of UNet inspired architectures for segmentation of cardiac structures such as the left and right ventricular cavity (LVC, RVC) and the left ventricular myocardium (LVM) on each time instance of the cardiac cycle. For the classification task, information is extracted from the segmented time-series in form of comprehensive features handcrafted to reflect diagnostic clinical procedures. Based on these features we train an ensemble of heavily regularized multilayer perceptrons (MLP) and a random forest classifier to predict the pathologic target class. We evaluated our method on the ACDC dataset (4 pathology groups, 1 healthy group) and achieve dice scores of 0.945 (LVC), 0.908 (RVC) and 0.905 (LVM) in a cross-validation over the training set (100 cases) and 0.950 (LVC), 0.923 (RVC) and 0.911 (LVM) on the test set (50 cases). We report a classification accuracy of 94% on a training set cross-validation and 92% on the test set. Our results underpin the potential of machine learning methods for accurate, fast and reproducible segmentation and computer-assisted diagnosis (CAD).

**Keywords:** Automated Cardiac Diagnosis Challenge, Cardiac Magnetic Resonance Imaging, Disease Prediction, Deep Learning, CNN

## 1 Introduction

Cardiac remodeling plays an inherent role in the progressive course of heart failure. The process results in poor prognosis for the patient due to diminished contractile systolic function, reduced stroke volume or malignant arrhythmia. Clinical manifestations are changes in size, mass, geometry, regional wall motion and function of the heart [1], which can be assessed timely and monitored non-invasively by cardiac magnetic resonance imaging (CMRI). In today's clinical routine, the huge benefits of comprehensive quantitative measurements are still not exploited due to the associated labour time, subjective biases and lack of reproducibility. Accurate automatic approaches for simultaneous multi-structure segmentation and CAD are thus desirable assets for a large spectrum of cardiac diseases.

---

⋆ contributed equally

Convolutional neural networks (CNN) have recently shown outstanding performance in medical image segmentation [2], where they typically take the form of UNet-like architectures [3]. So far, a limited number of fully-automatic cardiac CNN segmentation methods have been proposed [4–6], however, they did not consider to segment all volumes of the cardiac cycle. Highly accurate multi-structural segmentations on the entire cardiac cycle are crucial for automatic pathology assessment, especially when considering geometrical and dynamical changes of anatomical structures.

Computer-assisted diagnosis (CAD) approaches originate from the field of lesion detection and classification [7], which primarily focuses on texture information to discriminate healthy from pathological tissue. Medrano-Gracia et al. investigated global shape variations of the left ventricle in a large cohort of an asymptomatic population [8]. They found the major principal modes of shape variation to be associated with known clinical indices of adverse remodelling, including heart size, sphericity and concentricity. Later, Zhang et al. used a supervised method to extract the most discriminatory global shape changes associated with remodeling after myocardial infarction [9]. The resulting shape model was able to discriminate patients from asymptomatic subjects with 95% accuracy. However, to the best of our knowledge, a comprehensive CAD system for different cardiac remodelling pathologies and myocardial infarct patients has not been proposed before.

In this paper we present an approach for automatic classification of cardiac diseases associated with pathological remodelling. Based on multi-structure segmentation for each time step of the CMRI, we extract domain-specific features, which are motivated by a cardiologist's workflow, to then train an ensemble of classifiers for disease prediction (see Figure 1). We evaluated our methods for segmentation *and* classification on the MICCAI ACDC data set [10].

## 2 Methods

### 2.1 Cardiac cine-MRI Dataset

The ACDC dataset [10] comprises short-axis cine-MRI of 150 patients acquired at the University Hospital of Dijon using two MR scanners of different magnetic strengths (1.5 T and 3.0 T). Each time-series is composed of 28 to 40 3D volumes, which partially or completely cover the cardiac cycle. As typical for CMRI, the data is characterized by a high in-plane resolution ranging from 0.49 to 3.69 mm$^2$ and a low resolution in the direction of the long axis of the heart ($5 - 10$ mm slice thickness). Note that some data exhibit severe slice misalignments, which originate from different breath hold positions between slice stack acquisitions. Structures of interest, namely LVC, LVM and RVC were segmented manually by clinical experts on end diastolic (ED) and end sstolic (ES) phase instants. Four pathological groups and one group of healthy patients are evenly distributed in the dataset: patients with previous myocardial infarction (MINF), dilated cardiomyopathy (DCM), hypertrophic cardiomyopathy (HCM), abnormal right ventricle (ARV) and normal (healthy) subjects (NOR). Additional information for all patients is provided in form of height and weight. The dataset is split into 100 training and 50 test patients. Segmentation and classification ground truth is provided only for the 100 training cases. All reported test set results were obtained by submitting our predictions to the online evaluation platform.

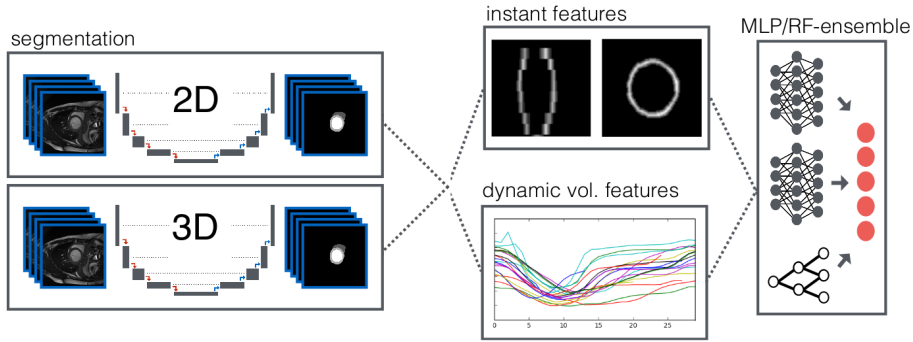

Fig. 1: Overview of the proposed pipeline: Segmentation predictions from a 2D and a 3D model are averaged and used to extract instant and dynamic volume features, which are fed into an ensemble of classifiers for disease prediction.

### 2.2 Segmentation

**Data preprocessing** For the segmentation part of the ACDC challenge we resampled all volumes to $1.25 \times 1.25 \times 10$ mm per voxel (for 3D UNet and feature extraction) and $1.25 \times 1.25 \times Z_{orig}$ mm per voxel (for 2D UNet) to account for varying spatial resolutions. The grey level information of every image was normalized to zero-mean and unit-variance.

**Network architecture** We tackle the segmentation using an ensemble of modified 2D and 3D UNets [3, 11]. The 3D segmentation model consists of a context aggregating pathway followed by a localization pathway. Both are interconnected at various scales to allow for recombination of abstract context features with the corresponding local information. We carefully adapted the architecture to cope with specific challenges of CMRI (see Figure 2): Due to low z-resolution of the input, pooling and upscaling operations are carried out only in the x-y-plane. Context in the z-dimension is solely aggregated through the 3D convolutions. Each feature extraction block (shown in gray) consists of two padded $3 \times 3 \times 3$ convolutions, followed by batch normalization and a leaky ReLU nonlinearity. Due to the shallow nature of the network (18 layers) no residual connections are utilized. The initial number of 26 feature maps is doubled (halved) with each of the 4 pooling (upscaling) operations, resulting in a maximum of 416 feature maps at the bottom of the U-shape. Deep supervision (as in [12]) is implemented by generating low resolution segmentation outputs via $1 \times 1 \times 1$ convolutions before each of the last two upscaling operations, which are upscaled and aggregated for the final segmentation.

The 3D model was trained for 300 epochs in a 5-fold cross validation using the ADAM solver and a pixel-wise categorical cross-entropy loss. The initial learning rate of $5 \cdot 10^{-4}$ was decayed by 0.98 per epoch, where an epoch was defined as 100 batches, each comprising four training examples. Training examples were generated as random crops of size $224 \times 224 \times 10$ voxels taken from a randomly chosen training patient and phase instance (ED/ES).

The 2D model's architecture is equivalent to the 3D approach except 2D convolutions. Due to the lower memory requirements, we increased the number of initial feature maps to 48. The network is trained with a batch size of 10 and input patches of size $352 \times 352$ pixels using a multiclass dice loss:

$$\mathcal{L}_{\text{dc}} = -\frac{2}{|K|} \sum_{k \in K} \frac{\sum_i u_i^k v_i^k}{\sum_i u_i^k + \sum_i v_i^k}, \quad (1)$$

where $u$ is the softmax output of the network and $v$ denotes a one hot encoding of the ground truth segmentation map. Both $u$ and $v$ are of size $i \times k$ with $i$ being the number of pixels in the training patch and $k \in K$ being the classes.

To accomplish the training of a well generalizing model on limited data, we used a broad range of data augmentation techniques, such as mirroring along the x and y axes, random rotations, gamma-correction and elastic deformations. Due to the low z-resolution all data augmentation was performed only in the x-y-plane. To account for the presence of slice misalignments, we artificially increased the number of misaligned slices by motion augmentation for the training of the 3D model: All slices within the training batch were perturbed with a probability of 10% and a random offset drawn from $\mathcal{N}(0, 20)$.

To obtain the final segmentations, softmax outputs of both networks were resampled to the original voxel resolution of the input image and then averaged.

### 2.3 Cardiac Disease Classification

**Feature Extraction** We extract two sets of features from the previously segmented structures to perform disease classification. All features were designed to quantify the traditional assessment procedures of expert cardiologists by describing static and dynamic properties of the structures of interest (see Table 1).

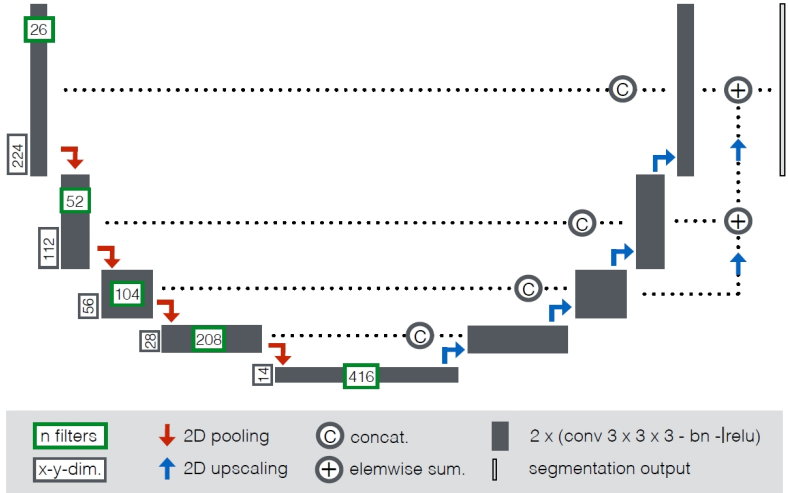

Fig. 2: Architecture of the 3D segmentation network. The 2D network is equivalent, but uses 2D convolutions, patch size $352 \times 352$ and 48 initial features.

*Instant features* Extracted from the two labeled ED and ES time instants as provided by the ACDC dataset, these features cover local and global shape information (circumference, circularity, LVM thickness, etc.), local variations (size of RVC at the apex, LVM thickness between RVC and LVC), simple texture descriptors (mass) as well as additional meta information (body mass index, weight, height). Notably, all thicknesses, circumferences and circularities are computed on the individual x-y-planes and aggregated over the z-dimension. The body surface is estimated from weight and height using the Mosteller formula.

*Dynamic volume features* We deployed the trained segmentation model to predict the anatomical structures in all time steps of the CMRI. This allows for exploitation of volume dynamics throughout the entire cardiac cycle independent of the predefined ED/ES. These volume dynamics are quantified in form of first order statistics (median, standard deviation, kurtosis, skewness) complemented by characteristics of the cardiac cycle's minimum and maximum volumes: We found the time instants of these extrema to not match the predefined ED/ES instants in the majority of patients. This finding is accounted for by computing volume, volume ratios and ejection fractions based on the determined actual minimum ($v_{\min}$) and maximum ($v_{\max}$) volume of the cardiac cycle. Finally, the synchrony of contraction between LVC and RVC is measured in form of the time step differences between their corresponding $v_{min}$ and $v_{max}$.

| instant features | RVC | LVM | LVC |
|---|---|---|---|
| max thickness* | | x | |
| min thickness* | | x | |
| std thickness* | | x | |
| mean thickness* | | x | |
| std thickness of LVM between LVC and RVC* | | | |
| mean thickness of LVM between LVC and RVC* | | | |
| mean circularity* | x | x | |
| max circumference* | x | x | |
| mean circumference* | x | x | |
| RVC size at most apical LVM slice* | | | |
| RVC to LVC size ratio at most apical LVM slice* | | | |
| volume per $m^2$ body surface | x | x | x |
| mass | | x | |
| patient weight | | | |
| patient height | | | |
| patient body mass index | | | |
| **dynamic volume features** | **RVC** | **LVM** | **LVC** |
| $v_{max}$ | x | x | x |
| $v_{min}$ | x | x** | x |
| dynamic ejection fraction | x | x** | x |
| volume median | x | x | x |
| volume kurtosis | x | x | x |
| volume skewness | x | x | x |
| volume standard deviation | x | x | x |
| volume ratio $v_{min,LVC}/v_{min,RVC}$ | | | |
| volume ratio $v_{min,LVM}/v_{min,LVC}$ | | | |
| volume ratio $v_{min,RVC}/v_{min,LVM}$ | | | |
| time step difference $t(v_{min,LVC})-t(v_{min,RVC})$ | | | |
| time step difference $t(v_{max,LVC})-t(v_{max,RVC})$ | | | |

Table 1: The two sets of features extracted for disease classification and the corresponding cardiovascular structure (RVC, LVM, LVC). All instant features (except for additional patient information) are extracted on both ED and ES.
*this feature was calculated in the x-y-plane and aggregated over slices in z.
** $v_{min,LVM}$ was determined at $t(v_{min,LVC})$.

**Classification** The features described in section 2.3 were used to train an ensemble of 50 multilayer perceptrons (MLP) and a random forest for pathology classification. The MLP's architecture consists of four hidden layers, each containing 32 units, followed by batch normalization, leaky ReLU nonlinearity and a Gaussian noise layer ($\sigma = 0.1$). Each MLP was trained on a random subset of 75% of the training data, while the remaining 25% were used for epoch selection. Further regularization was provoked by only presenting a random subset of 2/3 of the features to each MLP. We trained all MLPs for 400 epochs (with a patience of 40 epochs) using the ADAM solver with an initial learning rate of $5 \cdot 10^{-4}$, decayed by 0.97 per epoch. An epoch was defined as a set of 50 batches containing 20 patients each. Additionally, we trained a random forest with 1000 trees. During testing, the softmax outputs of all MLPs were averaged to obtain an overall MLP score, which was recombined subsequently with the random forest output to obtain the final ensemble prediction.

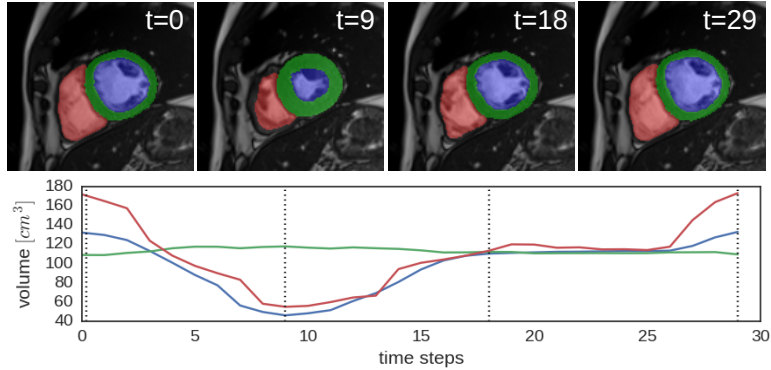

Fig. 3: Time-series segmentation for RVC (red), LVM (green), LVC (blue) and their corresponding volume dynamics. The example shows the central slice in z direction of a healthy patient (NOR).

## 3  Results

**Segmentation** With regard to the expert segmentations on the original ED and ES phase instants, individual dice scores of 0.945 for the LVC, 0.905 for the LVM and 0.908 for the RVC were achieved in 5-fold cross-validation (see Table 2 for detailed results including Hausdorff distances). When comparing performances of the 2D and 3D approach, the 3D model was largely outperformed by the 2D model (see Table 3). A marginal increase in RVC dice was observed when ensembling the models. Note that cardiac phase instances other than ED and ES were not considered in the scores due to unavailable ground truth labels. Qualitatively, the 4D segmentation yielded convincing results, which were smooth and robust in time for all substructures. CMRI with slice misalignments were segmented successfully by the model when using motion augmentation (3D network only). Based on the cross-validation, we observed only little overfitting, most of which occurred for the RVC region. The main mode of failure was the basal part of the RVC region, where the model struggled to distinguish between the right atrium and the right outflow tract or the RVC. This occurred mostly in ES images, resulting in a lower average dice score compared to ED. Other failures occurred in the LVC region, where papillary muscles were anatomical correctly classified as LVM but should have been classified as LVC to meet the convention of the challenge. This was especially observed in HCM patients. Simultaneous segmentation of multiple structures in a 3D volume of size $320 \times 320 \times 10$ voxels took less than one second for the 3D model and 1-2 seconds for the 2D model on a Pascal Titan X GPU.

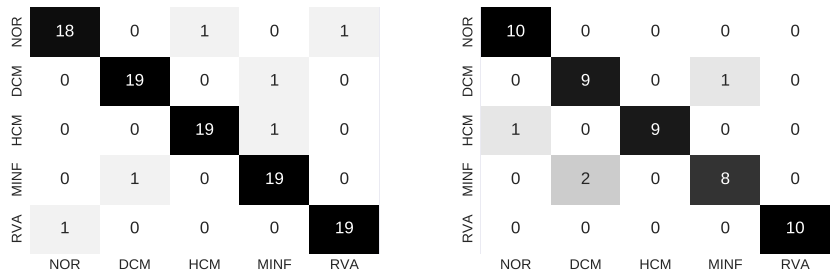

Fig. 4: Confusion matrices of the ensemble predictions from cross-validation on the training set (left) and on the test set (right). Rows correspond to the predicted class and columns to the target class, respectively.

**Classification** We trained the classification ensemble (see Section 2.3) on the ACDC training data using the features described in section 2.3. In a five fold cross-validation, a classification accuracy of 94% was achieved. The individual performance of the MLP ensemble and random forest were 93% and 92%, respectively. The test set accuracy was 92%. Confusion matrices are provided in Figure 4, indicating equal perfomance among classes in the cross-validation, and difficulties in distinguishing DCM from MINF patients on the test set. Feature computation took 15 s for instant features and less than one second for the dynamic volume features.

## 4  Discussion

In this paper we presented a fully automatic processing pipeline for pathology classification on cardiac cine-MRI. First, we developed an accurate multi-structure segmentation method trained solely on ED and ES phase instances, but capable of processing the entire cardiac cycle. Our approach revolves around the use of both a 2D and 3D model, leveraging their respective advantages through ensembling. The resulting pipeline is robust against slice misalignments, different CMRI protocols as well as various pathologies. We achieve dice scores of 0.950 (LVC), 0.923 (RVC) and 0.911 (LVM) on the ACDC test set, which earned us the first place in the segmentation part of the challenge. Based on the segmentations generated by our model, geometrical features are extracted and utilized by an ensemble of classifiers to predict the diagnosis, yielding promising outcomes. We ranked second in the classification part of the challenge with an accuracy of 92%. Our fully automatic processing pipeline constitutes an attractive software for clinical decision support due to the visualization of intermediate segmentation maps, the comprehensive quantification of cardiologic assessment and the rapid processing speed of less than 40 s. Possible future improvements of the model concern data augmentation and the architecture of the segmentation network as well as a regularization objective as used in [5]. Training the pipeline end-to-end in a multitask architecture could yield further improvement.

|   | Instance | Dice | | | Hausdorff (mm) | | |
|---|---|---|---|---|---|---|---|
|   |   | RVC | LVM | LVC | RVC | LVM | LVC |
| DCM | ED | 0.942 | 0.906 | 0.968 | 20.87 | 8.21 | 7.117 |
|  | ES | 0.872 | 0.913 | 0.916 | 17.944 | 8.161 | 5.886 |
|  | total | **0.907** | **0.910** | **0.942** | **19.830** | **8.153** | **6.544** |
| HCM | ED | 0.938 | 0.901 | 0.968 | 12.721 | 8.709 | 7.256 |
|  | ES | 0.878 | 0.907 | 0.935 | 18.326 | 11.355 | 14.514 |
|  | total | **0.908** | **0.904** | **0.952** | **15.321** | **10.105** | **11.022** |
| MINF | ED | 0.937 | 0.896 | 0.961 | 13.385 | 9.63 | 6.882 |
|  | ES | 0.889 | 0.907 | 0.907 | 18.639 | 11.74 | 9.599 |
|  | total | **0.913** | **0.901** | **0.934** | **16.107** | **10.730** | **8.116** |
| NOR | ED | 0.939 | 0.887 | 0.971 | 9.765 | 7.231 | 4.626 |
|  | ES | 0.884 | 0.901 | 0.941 | 11.407 | 9.164 | 7.665 |
|  | total | **0.911** | **0.898** | **0.956** | **10.615** | **8.397** | **6.330** |
| RV | ED | 0.948 | 0.909 | 0.964 | 14.728 | 10.716 | 9.392 |
|  | ES | 0.852 | 0.911 | 0.922 | 15.126 | 11.769 | 11.224 |
|  | total | **0.900** | **0.91** | **0.943** | **15.133** | **11.605** | **10.620** |

Table 2: Dice scores and Hausdorff distances of the segmentation model for pathological subgroups (results of 5-fold cross validation).

|   |   | Dice | | | Hausdorff (mm) | | |
|---|---|---|---|---|---|---|---|
|   |   | RVC | LVM | LVC | RVC | LVM | LVC |
| CV | 2D model | 0.902 | 0.905 | 0.945 | 14.294 | 8.899 | 7.055 |
|  | 3D model | 0.879 | 0.872 | 0.928 | 16.289 | 10.438 | 9.778 |
|  | ensemble | **0.908** | **0.905** | **0.945** | **15.291** | **9.668** | **8.416** |
| test | ensemble | **0.923** | **0.911** | **0.950** | **11.134** | **8.696** | **7.145** |

Table 3: Dice scores and Hausdorff distances of the segmentation. Results from cross-validation (CV) on the training set are shown for the 2D model, the 3D model and the corresponding ensemble. On the test set, only ensemble results are shown.

## Acknowledgements

The author Sandy Engelhardt was funded by the German Research Foundation (DFG) as part of project B01, SFB/TRR 125 Cognition-Guided Surgery.